\documentclass[runningheads]{llncs}
\usepackage{graphicx}

\usepackage{tikz}
\usepackage{comment}
\usepackage{amsmath,amssymb} 
\usepackage{color}
\usepackage{subcaption}
\newcommand{\mbf}[1]{\mathbf{#1}}
\newcommand{\etal}{\textit{et al}. }

\begin{document}
\pagestyle{headings}
\mainmatter
\def\ECCVSubNumber{100}  

\title{maskGRU: Tracking Small Objects in the Presence of Large Background Motions}

\author{Constantine J. Roros, Avinash C. Kak}
\institute{Purdue University}
\titlerunning{Tracking Small Objects in the Presence of Large Background Motion}

\maketitle

\begin{abstract}
We propose a recurrent neural network-based spatio-temporal framework named maskGRU for the detection and tracking of small objects in videos. While there have been many developments in the area of object tracking in recent years, tracking a small moving object amid other moving objects and actors (such as a ball amid moving players in sports footage) continues to be a difficult task. Existing spatio-temporal networks, such as convolutional Gated Recurrent Units (convGRUs) \cite{BallasDelve}, are difficult to train and have trouble accurately tracking small objects under such conditions.
To overcome these difficulties, we developed the maskGRU framework that uses a weighted sum of the internal hidden state produced by a convGRU and a 3-channel mask of the tracked object’s predicted bounding box as the hidden state to be used at the next time step of the underlying convGRU.
We believe the technique of incorporating a mask into the hidden state through a weighted sum 
has two benefits: controlling the effect of exploding gradients and introducing an attention-like mechanism into the network by indicating where in the previous video frame the object is located.
Our experiments show that maskGRU outperforms convGRU at tracking objects that are small relative to the video resolution even in the presence of other moving objects.
\keywords{Recurrent neural networks, Deep learning, Object tracking}
\end{abstract}

\section{Introduction}
\begin{figure}[h!]
\begin{subfigure}{\textwidth}
  \centering
  \includegraphics[width=\linewidth]{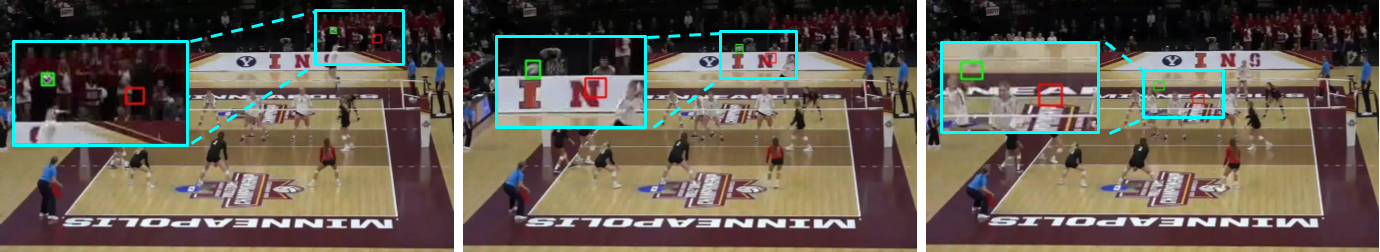}
  \caption{convGRU}
  \label{fig:convgru_sfig}
\end{subfigure}\\
\begin{subfigure}{\textwidth}
  \centering
  \includegraphics[width=\linewidth]{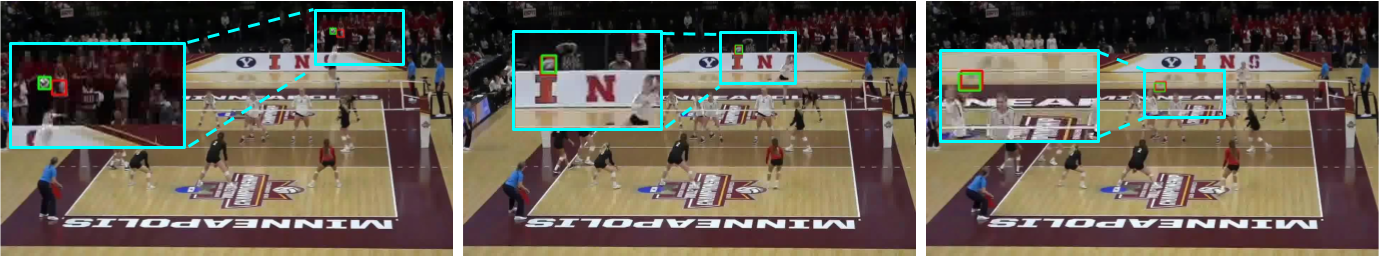}
  \caption{maskGRU}
  \label{fig:maskgru_sfig}
\end{subfigure}%
\caption{A sampling of the results obtained with maskGRU and a convGRU network on volleyball video frames over the span of 1 second. The green bounding box is the ground truth for the volleyball and the red bounding box is the one predicted by the corresponding network. Due to a combination of motion blur and low resolution, the volleyball is not discernible to the human eye in the rightmost frames.
}
\label{fig:maskgru_v_convgru}
\end{figure}
Object tracking is a problem that has received considerable attention in the field of computer vision. 
Some approaches simply use an object detector framework to find an object in each frame of a video, ignoring any information in previous frames.
But when trying to track a small object in motion, it is useful to leverage temporal information in the video to aid in following the object. This is especially true when one wants to track a moving object in the presence of other actors or objects in motion. Trying to track a volleyball in sports footage, for example, is difficult due to the volleyball being small relative to the size of the playing field and the motion of the players adding complexity to the scene.

Recurrent neural networks (RNNs) such as Gate Recurrent Units (GRUs) \cite{cho2014} and Long Short-Term Memory networks (LSTMs) \cite{hochreiter1997} are frequently used in temporal applications due to their ability to leverage previous information in a time series to make a prediction at the current time step. Since videos are not just temporal but spatial in nature, it is appropriate to use the convolutional RNNs such as convGRU \cite{BallasDelve} or convLSTM \cite{shi2015} that replace the fully-connected layers of the original GRU and LSTM with convolutional layers to process spatio-temporal data. Therefore, using a convGRU for the purpose of tracking a moving object seems like a logical conclusion.

In practice, however, convGRUs face several difficulties in tracking a moving object. Despite the use of a gating mechanism  to help combat vanishing gradients, convGRUs are still susceptible to this problem, making training difficult. Techniques such as teacher forcing can help guide network training, but can result in a network that lacks robustness. For example, a convGRU trained with teacher forcing may follow the trajectory of an object but with an offset from the true location of the object, giving the appearance of a ``time delay" in the tracker (an example of which is shown in Figure \ref{fig:convgru_sfig}). 
Alternatives to GRU such as STAR \cite{turkoglu2021Star} have been developed to  mitigate the effects of exploding gradients, but lack the capacity to perform a task as difficult as tracking small objects in the presence of large background motions.

We introduce the maskGRU framework that uses a modified hidden state feedback loop to improve small-object tracking performance. Typically in an RNN, there is a feedback loop  where the output hidden state from the previous time step is fed as input into the next time step. maskGRU instead uses a weighted sum of the output hidden state and a 3-channel mask as input into the next time step, where the mask is produced using the bounding box prediction at the current time step. The bounding box prediction itself is produced by feeding the output hidden state into a neural network with convolutional and fully-connected layers. By weighting the hidden state with this mask, the effects of vanishing gradients are dampened. Furthermore, we believe the bounding box mask, when combined with hidden state, 
acts as an attention map in that it helps indicate which area of the image the network should focus on.

The rest of the paper is outlined as follows. Section \ref{sec:related} discusses previous work on architectures for tracking small objects and prior spatio-temporal recurrent architectures. Section \ref{sec:maskgru} describes the maskGRU framework and discusses potential reasons for its effectiveness. In section \ref{sec:experiments} we discuss the experiments performed and the consequent results.

\section{Related Work}
\label{sec:related}
In this section we will first discuss RNNs  that are designed to process spatio-temporal data such as videos. While not all such RNNs are designed with object tracking in mind, such architectures can potentially be used in an object tracking framework. Then we will discuss frameworks specifically for tracking small objects.
\subsection{Spatio-Temporal RNNs}
\label{sec:relatedrnn}
Convolutional RNNs that process data both over space and time have been frequently used not just for object tracking, but for other applications such as action recognition.
Ballas \etal was one of the earliest papers that used convolutional layers in a GRU  in what they call a GRU-Recurrent Convolutional Network (GRU-RCN) \cite{BallasDelve}. GRU-RCN does not use images directly as input, uses feature maps produced by VGG-16 \cite{simonyan2015}. Ballas \etal experimented with different configurations of the GRU-RCN and found that stacked, bi-directional convGRU units worked the best for action recognition. Around the same time, Shi \etal developed a convLSTM \cite{shi2015} to predict weather using radar maps using a similar method but with LSTMs instead of GRUs.

Li \etal uses two convLSTMs simultaneously for action recognition \cite{LiVideoLSTM}. The first convLSTM produces a soft attention map of the for the video frames using optical flow images as input. This attention map is applied to feature maps produced from RGB image frames that were fed into VGG. These feature maps with attention applied are then fed into the second convLSTM, whose output is then fed into a fully-connected layer to produce an action prediction.

Tokmakov \etal uses RGB images and optical flow images simultaneously  to produce semantic segmentation of videos \cite{TokmakovVisualMemSeg}. Each image type is fed into a sub-network that extracts visual and motion feature maps. These feature maps are concatenated and fed into a bi-directional convGRU, whose output is then fed into a convolutional layer to provide the final segmentation output.

Turkoglu \etal introduces the STAR unit in order to construct deep (i.e. multiple stacked recurrent units) RNNs \cite{turkoglu2021Star}. Of particular interest to us in this work are experiments with a convolutional variant of STAR for the purpose of gesture recognition. The desire for deeper RNNs comes from the general notion that deeper neural network architectures generally do a better job at extracting features. But (as the authors show) RNNs are particularly susceptible to vanishing or exploding gradients, and therefore architectures of stacked recurrent units usually do not perform well. 
Therefore, STAR is designed ensure that gradients back-propagate in a stable manner. On top of this, STAR aims to reduce the number of parameters relative to GRU or LSTM. 
\subsection{Tracking Small Objects}
\label{sec:relatedtracking}
There have been several attempts at developing a tracking method specifically for small moving objects. The TrackNet network \cite{huang2019} uses an encoder-decoder architecture to  predict the location of a ball or shuttlecock using consecutive frames of a tennis or badminton game. Three consecutive frames are concatenated channel-wise and fed into a VGG-like encoder to obtain feature maps of the frames, then uses DeconvNet \cite{noh2015} to produce a ``probability-like heatmap" (modeled as a scaled 2D Gaussian distribution located at the center of the object for the ground-truth during training) of the object location in the most recent input frame. TrackNetv2 \cite{sun2020} extends TrackNet to produce multiple heatmaps for each input frame and incorporates U-Net-like skip connections \cite{ronneberger2015}, among other changes.

Marvasti-Zadeh \etal takes inspiration from the ATOM (Accurate Tracking by Overlap Maximization)  \cite{danelljan2019} object tracker to create the COMET (Context-aware IoU-guided network for small object tracking)  \cite{marvastizadeh2020} framework  specifically for tracking small objects in aerial imagery. Using a single reference frame annotated with the bounding box of an object, the network estimates the location of the object in future ``target" frames. COMET does this by first estimating the center of the object in the first target frame using ATOM’s fully-convolutional classifier that distinguishes the object from the background by outputting a 2D confidence score map. Using this predicted object center and the previous frame’s bounding box size, an initial bounding box is estimated for the target frame. A set of  candidate bounding boxes are produced by adding random Gaussian noise to these initial estimated bounding box coordinates.  These bounding box candidates are used to pool features from the target frame feature maps. The features are channel-wise multiplied by ``modulation vectors" in order to incorporate information about the object’s appearance into the target frame features.  The modulation vectors are created by feeding the annotated reference image into a network consisting of convolutional, precise ROI pooling \cite{jiang2018}, and fully-connected layers . Specifically, Inception V3 \cite{szegedy2016} is used to extract multi-scale features and bottleneck attention  modules \cite{park2018bam} are used to attend to relevant information in the frame feature maps. For each bounding box candidate’s respective feature map, the network predicts the intersection over union (IoU) of the candidate with respect to the target object. The final IoU prediction is taken as the average of the top bounding box candidates with the highest IoU.

Canepa \etal proposes  T-RexNet \cite{canepa2021} for tracking small moving objects in a hardware-efficient manner. Three consecutive video frames are processed into two images, one containing purely motion information and the other a mix of motion and visual features (where the motion features are obtained via image differencing). Features maps are obtained from each image using CNNs and are concatenated, then fed into SSD \cite{liu2016ssd} to obtain object bounding boxes and class confidence scores.

While the spatio-temporal RNNs mentioned in  section \ref{sec:relatedrnn} show promising results in action recognition, it is not clear how well they would be able to track small objects in the presence of large background motions. maskGRU is able to track such objects much better than the networks we have compared it with so far (convGRU and convSTAR). Some object tracking frameworks such as TrackNetv2 and T-RexNet have already been shown to perform small-object tracking in tennis and badminton videos. However, since there are fewer players on the court in tennis and badminton, such sports have less complex background scenes compared to volleyball, where six players are present on each side of the court. With more complex background scenery, the task of tracking a volleyball in gameplay footage is more challenging.
\section{The maskGRU Framework}
\begin{figure}[t]
    \centering
    \includegraphics[width=\linewidth]{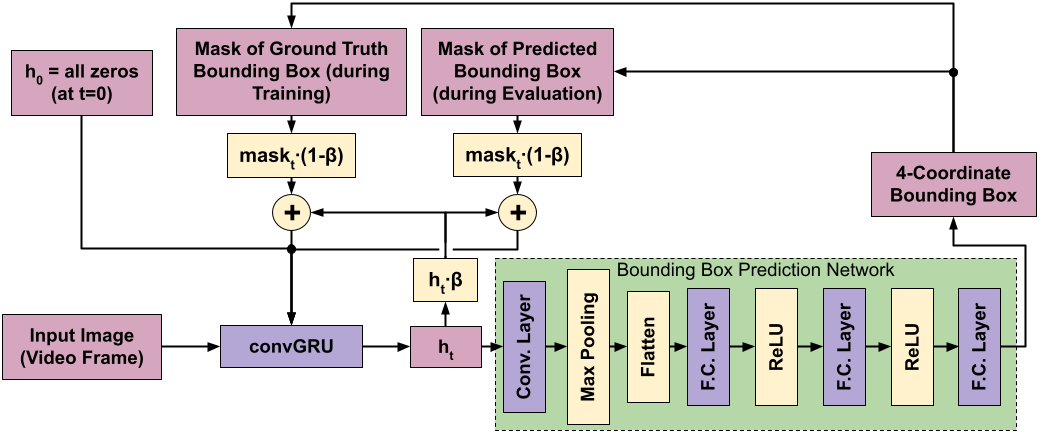}
    \caption{Overview of the maskGRU framework}
    \label{fig:maskgru}
\end{figure}
\label{sec:maskgru}
To understand the operation of maskGRU, we first start with the equations for a convGRU with input $\mbf{x}_t$ at time step $t$:
\begin{equation}
\label{eq:rt_convgru}
\mbf{r}_t = \sigma(\mbf{\mbf{W}_\mbf{r}}*[\mbf{h}_{t-1},\mbf{x}_t]+\mbf{b}_\mbf{r})
\end{equation}
\begin{equation}
\mbf{z}_t = \sigma(\mbf{\mbf{W}_z}*[\mbf{h}_{t-1},\mbf{x}_t]+\mbf{b}_{\mbf{z}})
\end{equation}
\begin{equation}
\hat{\mbf{h}}_t = \tanh(\mbf{W}_{\hat{\mbf{h}}}*[\mbf{h}_{t-1}\odot\mbf{r}_t,\mbf{x}_t]+\mbf{b}_{\mbf{\hat{h}}})
\end{equation}
\begin{equation}
\label{eq:ht_convgru}
\mbf{h}_t = \mbf{h}_{t-1}\odot(\mbf{1}-\mbf{z}_t)+\mbf{z}_t\odot\hat{\mbf{h}}_t
\end{equation}
\noindent
where $\odot$ is the element-wise product and
$[\mbf{a},\mbf{b}]$ indicates the channel-wise concatenation of tensors
$\mbf{a}$ and $\mbf{b}$. Here $\mbf{x}_t$ is a tensor of shape $3 \times W \times H$
(assuming it is an RGB image). 
We assume $\mbf{z_t}$,
$\mbf{r_t}$, and $\mbf{h_t}$ also have axes with these
dimensions. For each convolutional filter $\mbf{W}$ of size $k \times k$
in the convGRU, a padding of $\frac{k-1}{2}$ ensures that
each convolution layer output is the same size as the input
(assuming $k$ is odd).

Before we show how this network is modified into maskGRU, we first need to show how the output hidden state $\mbf{h}_t$ can be used to make a bounding box prediction $bb_t$. This can be done using the following bounding box prediction network:
\begin{equation}
\label{eq:hpool}
    \mbf{h}_{pool_t} = \mbf{W}_{pool} * MaxPool\left(\mbf{h}_t \right)+ \mbf{b}_{pool}
\end{equation}
\begin{equation}
    \label{eq:bb}
    bb_t =  \mbf{W}_{bb} \cdot NN_2\left(NN_1\left(Flatten\left(\mbf{h}_{pool_t}\right)\right)\right) + \mbf{b}_{bb}
\end{equation}
where  
\begin{equation}
    NN_l(x)=ReLU\left(\mbf{W}_{NN_l}\cdot x + \mbf{b}_{NN_l}\right)
\end{equation}
$\mbf{h}_{pool_t}$ is a tensor of shape $1 \times W_{pool} \times H_{pool}$, and the $Flatten(\cdot)$ function reshapes $\mbf{h}_{pool_t}$ into  $1 \times (W_{pool} H_{pool})$.  $\mbf{W}_{pool}$ is a convolutional filter of size $k \times k$ while $\mbf{W}_{bb} $,$\mbf{W}_{NN_1}$ and $\mbf{W}_{NN_2}$ are fully-connected layer weights.

We can produce a 3-channel tensor of zeros with the same shape as $\mbf{x}_t$ with the predicted bounding box ``drawn" at its corresponding pixel location in the tensor. This bounding box is drawn by setting the values at the locations of the bounding box outline in one of the tensor channels to 255. This tensor forms a mask $mask_t$ of the predicted bounding box $bb_t$ at time step $t$. This mask can then be weighted by some value $\beta \in [0,1]$ and summed with $\mbf{h}_{t}$ to produce a modified hidden state to be used in future time steps:
\begin{equation}
    \mbf{h}_{mask_t} = \beta \mbf{h}_{t} + (1-\beta)mask_t
\end{equation}

Using this modified hidden state, we can modify  equations (\ref{eq:rt_convgru}) through (\ref{eq:ht_convgru}) for  convGRU to obtain the following equations for maskGRU:

\begin{equation}
\label{eq:rt_maskgru}
\mbf{r}_t = IN(\sigma(\mbf{\mbf{W}_\mbf{r}}*[\mbf{h}_{mask_{t-1}},\mbf{x}_t]+\mbf{b}_\mbf{r}))
\end{equation}
\begin{equation}
\mbf{z}_t = IN(\sigma(\mbf{\mbf{W}_z}*[\mbf{h}_{mask_{t-1}},\mbf{x}_t]+\mbf{b}_{\mbf{z}}))
\end{equation}
\begin{equation}
\hat{\mbf{h}}_t = IN(\tanh(\mbf{W}_{\hat{\mbf{h}}}*[\mbf{h}_{mask_{t-1}}\odot\mbf{r}_t,\mbf{x}_t]+\mbf{b}_{\mbf{\hat{h}}}))
\end{equation}
\begin{equation}
\label{eq:ht_maskgru}
\mbf{h}_t = \mbf{h}_{mask_{t-1}}\odot(\mbf{1}-\mbf{z}_t)+\mbf{z}_t\odot\hat{\mbf{h}}_t
\end{equation}
In addition to incorporating $\mbf{h}_{mask_{t-1}}$, we have also added instance normalization layers \cite{ulyanov2016} $IN(\cdot)$ into the network. Conventionally, RNNs do not use batch normalization as it does not consider how input changes over time and therefore each time step would need its own normalization parameters (though some works such as Cooijmans \etal \cite{cooijmans2016} have attempted to incorporate it). We chose to use instance normalization, which avoids this issue, after each activation function in maskGRU as it resulted in the loss converging to a lesser value during training. The process for predicting $bb_t$ from $\mbf{h}_t$ in maskGRU is the same as the one for convGRU shown in equations (\ref{eq:hpool}) and (\ref{eq:bb}). The network layout of the maskGRU framework is presented in Figure \ref{fig:maskgru}.
\begin{figure}[ht]
\begin{subfigure}{.45\textwidth}
  \centering
  \includegraphics[width=0.9\linewidth]{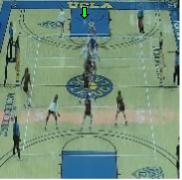}
  \caption{RGB frame input into maskGRU (We have used the green arrow to depict the location of the ball. The arrow is not in the image that is fed into the network)}
  \label{fig:sfig1}
\end{subfigure}\hspace{0.5cm}
\begin{subfigure}{.45\textwidth}
  \centering
  \includegraphics[width=0.9\linewidth]{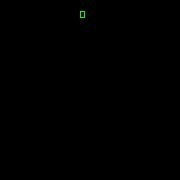}
  \caption{Mask used to create the input hidden state that is fed into maskGRU}
  \label{fig:sfig2}
  \vspace{0.75cm}
\end{subfigure}\\
\begin{subfigure}{.45\textwidth}
  \centering
  \includegraphics[width=0.9\linewidth]{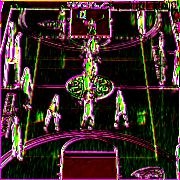}
  \caption{Output hidden state $\mbf{h}_t$}
  \label{fig:sfig3}
\end{subfigure}%
\begin{subfigure}{.6\textwidth}
  \centering
  \includegraphics[width=0.9\linewidth]{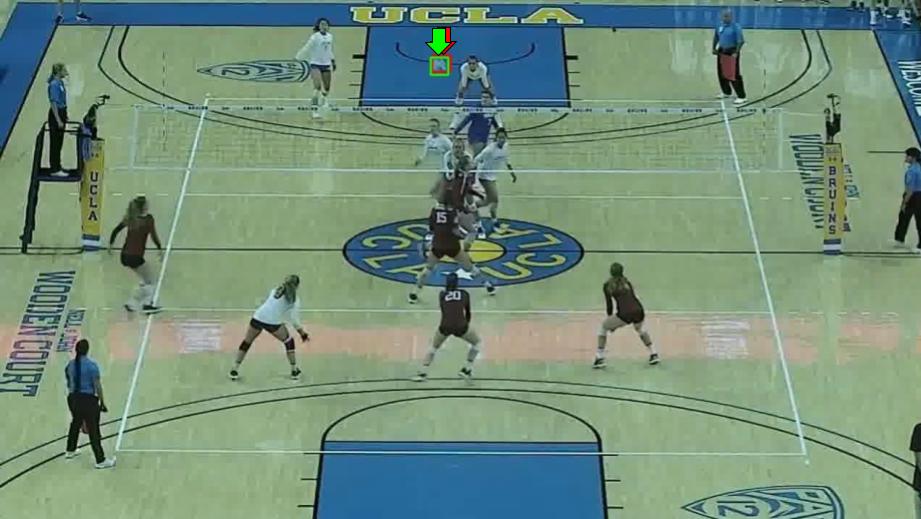}
  \caption{RGB frame (at original resolution) with the ground truth (green) and the predicted (red) bounding boxes}
  \label{fig:sfig4}
\end{subfigure}%
\caption{Visualization of maskGRU inputs and outputs for sample video frame}
\label{fig:visumaskgru}
\end{figure}
\subsection{Possible Reasons for maskGRU's Effectiveness}
We speculate that using a mask in the manner shown above results in improved object tracking performance by helping reduce the effects of vanishing gradients during training. Even though the gating mechanism of GRUs is designed to help mitigate the effects of vanishing gradients, the use of gates may not be sufficient to completely stop them. As we increase the value of $\beta$, we found that the value of the weights $\mbf{W}_{NN_2}$ in the second-to-last layer of the bounding box prediction network decreased. This layer precedes an ReLU layer, and when examining the  output of this ReLU for larger values of $\beta$, we see a dying ReLU effect with the layer output eventually becoming all zeros over the span of the sequence. Even in our experiments that used Parametric ReLU \cite{he2015delving}, which avoids the dying ReLU problem, we found that maskGRU still performs better when using a lower value of $\beta$, suggesting that the problem is deeper than simply a choice of the activation function. 

While it is true that these dying ReLU layers are at the end of the network, and the layers that directly suffer from the effects of vanishing gradients are at the beginning of the network, it is not unreasonable to interpret the ReLU output as a sign of vanishing gradients. This is because the output of the early network layers that are affected by vanishing gradients is forward propagated through the network, resulting in a ReLU input that is less than zero and consequently a ReLU output that is zero. Being at the end of the network, the dying ReLUs provide an easily observable and dramatic consequence of the vanishing gradients in the earlier stages of the network.

It is also possible that what helps make maskGRU effective is the role of the mask in indicating the region of interest within the input video frame. In a sense, it acts like an attention map by incorporating into the hidden state what spatial area of the frame is important. To illustrate, we show a sample video frame from a given time step in Figure \ref{fig:sfig1}. We see that the volleyball is hard to discern due to motion blur and the low resolution at which the frame is fed into the network. 
The output hidden state at the current time step is visualized in Figure \ref{fig:sfig3}. Here we can see that the output hidden state effectively extracts low-level features from both the input frame and the mask (shown in Figure \ref{fig:sfig2}). This output hidden state is fed into the bounding box prediction network to produce the bounding box prediction in Figure \ref{fig:sfig4}. Even though the bounding box mask is from the previous time step (and therefore probably would not be directly overlapping the ball in the current time step), we speculate that the mask still helps the network understand the general location of the ball as it is a prominent visual feature in the output hidden state.
\section{Experiments}
\label{sec:experiments}
To show the effectiveness of maskGRU, we compare it with convGRU for tracking the ball in volleyball gameplay videos. The ball is relatively small and in motion while the volleyball players are moving throughout the court. Videos were recorded with a camera at a fixed position and angle above the volleyball court, and with the ground truth bounding boxes manually annotated. Footage was recorded at 15 different courts, with anywhere from 2 to 20 videos per court, totalling 88 videos in all. The videos were originally captured at $1280 \times 720$ pixel resolution at 60 frames per second and are about 5 seconds long on average. 
\subsection{Training Procedure}
\label{sec:experimentsproc}
 For the sake of memory and computational efficiency, the video frames are resized to  $180 \times 180$ pixel resolution and the frame rate is downsampled to 30 frames per second.  To simplify training when using a batch size greater than 1, the sequence length of all samples is fixed at 60 frames (i.e. 2 seconds of video footage per sample), resulting in about 21,000 sequences total over all the videos.  Roughly 70\% of the videos are used for training while the rest are sequestered for validation.

The network was trained using the ADAM optimizer \cite{kingma2014adam} with a learning rate of 0.0001 and a batch size of 8. By default the network trains for 100 epochs, but training is stopped when either the training loss has converged 
or the validation loss starts to increase. The loss is computed as the smooth L1 loss \cite{girshick2015fast} between the predicted bounding box and the ground truth bounding box at each time step. 

When training maskGRU, we use a teacher forcing schedule. Teacher forcing is sometimes used in training RNNs, and involves replacing the hidden state of the RNN with the ground truth for use in the next time step, effectively ``forcing" the RNN to learn how the desired output is represented. Specifically in the case of maskGRU, teacher forcing is used when creating $\mbf{h}_{mask_t}$ by using a mask based on the ground truth bounding box instead of the predicted bounding box. A teacher forcing schedule is employed to allow the network to gradually adjust to using its own output instead of the ground truth over the course of training. At each training iteration, teacher forcing is used (i.e. the ground truth bounding box is used to produce $mask_t$) with probability $\epsilon=max(1-\gamma(epoch\ number),0)$, where $\gamma$ is a hyperparameter. In our experiments we found that setting $\gamma=0.01$ resulted in the best converged loss during training.

For comparison, we also trained convSTAR and convGRU on this dataset. To predict a bounding box, we incorporated maskGRU's bounding box prediction network into these models (but did not use the output bounding box to create a mask for use in future time steps). We tried to keep the training and model parameters for convSTAR and convGRU as close as possible to those of maskGRU. However, convGRU was particularly difficult to train, with extremely high average loss values and, as suggested by an extremely low ($10^{-8}$) average training IOU over multiple epochs, was unable to perform any semblance of tracking. We found that we could train a convGRU network in a stable manner if we removed the instance normalization layers from the convGRU, used a batch size of 1, used a shallower bounding box prediction network, and lowered $\gamma$ to a value of 0.005.

\subsection{Results}
\label{sec:experimentsresults}
\begin{table}[t]
\centering
\begin{tabular}{|l|l|l|}
\hline
         & Avg. Training IOU & Avg. Validation IOU \\ \hline
maskGRU  &   \textbf{0.3009}                &   \textbf{0.002531}  \\ \hline
convGRU  &   0.03092          &        0.002008    \\ \hline
convSTAR &     0.007945      &  0.001509           \\ \hline
\end{tabular}
\caption{Comparison of the average IOU for maskGRU, convGRU, and convSTAR on training and validation data}
\label{tab:results}
\end{table}
Results are shown in Table \ref{tab:results} for maskGRU, convGRU, and convSTAR models trained roughly for the same number of epochs. maskGRU has a much higher average training IOU on both training and validation data compared to both convGRU and convSTAR. The results also show how maskGRU struggles to track well on validation data. Nonetheless, it still performs better than convSTAR and convGRU on the validation set. Future work will explore how to improve maskGRU's ability to track unseen video sequences.

\section{Conclusion}
In this paper we present the maskGRU framework for tracking small objects in the presence of large background motions. Our network builds on the convGRU network by altering how the hidden state is used in subsequent time steps. By using a weighted sum of the output hidden state and a mask of the predicted bounding box, we are better able to track a volleyball in gameplay videos than with a normal convGRU. We speculate that incorporating a mask into the hidden state helps mitigate the effects of vanishing gradients and acts as an attention-like mechanism by indicating the region of interest in the image.
In future work we plan to further explore the reasons behind maskGRU's effectiveness through ablation experiments. We also plan to compare it with other models beyond convGRU such as COMET and TrackNetv2.


%
%
\bibliographystyle{splncs04}
\bibliography{egbib}
\end{document}